\documentclass{article}

\usepackage{microtype}
\usepackage{graphicx}
\usepackage{subfigure}
\usepackage{booktabs} % for professional tables
\usepackage{amsmath}
\usepackage{amssymb}
\usepackage{multirow}
\usepackage[table]{xcolor}
\usepackage{colortbl}
\usepackage{placeins}
\usepackage{needspace}
\DeclareRobustCommand{\revise}[1]{#1}
\DeclareRobustCommand{\dolphinicon}{\raisebox{-0.15em}{\includegraphics[height=1.1em]{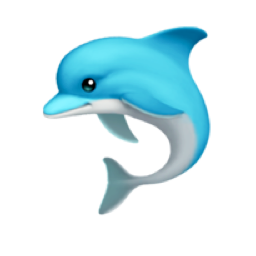}}\hspace{0.3em}}
\usepackage{hyperref}
\usepackage[accepted]{mlsys2025}
\hypersetup{allcolors=black,pdfsubject={Preprint}}
\makeatletter
\renewcommand{\Notice@String}{}
\makeatother

\AddToHook{cmd/subsection/before}{\Needspace{5\baselineskip}}
\AddToHook{cmd/subsubsection/before}{\Needspace{5\baselineskip}}

\makeatletter
\let\pdroute@float@makebox\float@makebox
\renewcommand{\float@makebox}[1]{%
  \color@vbox\normalcolor
  \pdroute@float@makebox{#1}%
  \color@endbox}
\makeatother

\mlsystitlerunning{\revise{DolphinRouting: Cluster-Level Binary Routing for LLM Serving}}

\begin{document}

\twocolumn[
\mlsystitle{\revise{\dolphinicon DolphinRouting:} Cluster-Level Binary Routing for Disaggregated LLM Inference Serving}

% Author layout inspired by https://cyberdual.github.io/cyberdual.pdf.
% Keep the existing MLSys title, abstract, and body style.
% Small vector emblems replace numeric affiliation markers.
\begingroup
% Apple Color Emoji glyphs (U+1F31F / U+1F4E7), embedded for pdfLaTeX.
% Keep the former superscript's height so the author block does not reflow.
\newcommand{\equalcontributionmark}{\hspace{0.12em}%
  \raisebox{0.05ex}[0pt][0pt]{%
    \includegraphics[height=0.85em]{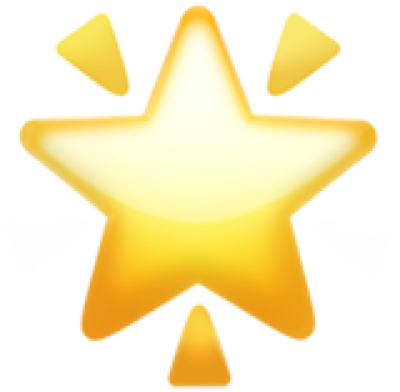}}%
  \vphantom{\textsuperscript{*}}}
\newcommand{\correspondingauthormark}{\hspace{0.12em}%
  \raisebox{\dimexpr0.4em+0.15ex-0.5\height\relax}[0pt][0pt]{%
    \includegraphics[width=0.85em,height=0.65em,keepaspectratio]{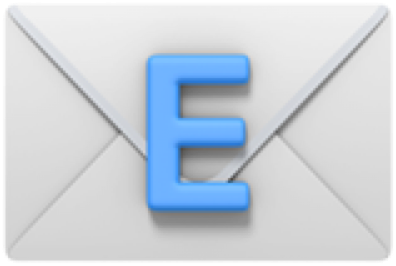}}%
  \vphantom{\textsuperscript{*}}}
\newcommand{\njuauthormark}{\,\raisebox{0.15ex}[0pt][0pt]{%
  \includegraphics[height=0.8em]{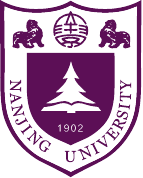}}}
\begin{mlsysauthorlist}
\vspace{-8pt}
\textbf{Yishuo Yuan}\njuauthormark\equalcontributionmark\hspace{1.2em}%
\textbf{Shaoyu Yang}\njuauthormark\equalcontributionmark\hspace{1.2em}%
\textbf{Chunrong Fang}\njuauthormark\correspondingauthormark
\par\vspace{1pt}
\raisebox{-0.15ex}{\includegraphics[height=0.9em]{figures/nju-emblem.pdf}}\hspace{0.35em}%
Nanjing University\qquad{\small\equalcontributionmark\,Equal contribution}
\par\vspace{1pt}
{\small\texttt{231870185@smail.nju.edu.cn}\quad
\texttt{shaoyuyoung@gmail.com}\quad
\correspondingauthormark\,\texttt{fangchunrong@nju.edu.cn}}
\end{mlsysauthorlist}
\endgroup
\hypersetup{pdfauthor={Yishuo Yuan, Shaoyu Yang, Chunrong Fang}}
\mlsyskeywords{Machine Learning, MLSys}

% Reuse the original author-area height so the abstract and body stay put.
\vskip 3.681pt

\begin{abstract}
Prefill-decode (PD) disaggregation isolates the two phases of large language model serving, but request-level path selection must account for prefill work, cache reuse, and decode demand. We present DolphinRouting, which learns this choice from end-to-end SLO feedback under a given P/D resource allocation. Requests are clustered by raw prompt length, cache-adjusted input lengths on both worker pools, and output-length information. Offline policy search learns one binary action per cluster. Online feature extraction, cluster assignment, and table lookup select local or remote prefill. We evaluate DolphinRouting on a real PD-disaggregated cluster using three public benchmarks, a mixed workload, and real data. Across the mixed workload and real data, DolphinRouting with $k$-means reduces P90 TTFT by 5.7--7.9\% and P90 TPOT by 2.6--10.1\% relative to Dynamo.
\end{abstract}
]

% Author footnote for the public version. Reserve the former notice's height
% as blank space to preserve the body positions and page breaks.
\makeatletter
{\let\thefootnote\relax\footnotetext{%
\hspace*{-\footnotesep}\vphantom{\textsuperscript{1}}Nanjing University, Nanjing, China.\newline
Correspondence to: Chunrong Fang\newline
\textless{}fangchunrong@nju.edu.cn\textgreater{}.\par
\vspace*{3\baselineskip}
}}
\makeatother

\section{Introduction}
\label{sec:intro}

\begin{figure}[t]
    \centering
    \includegraphics[width=\linewidth]{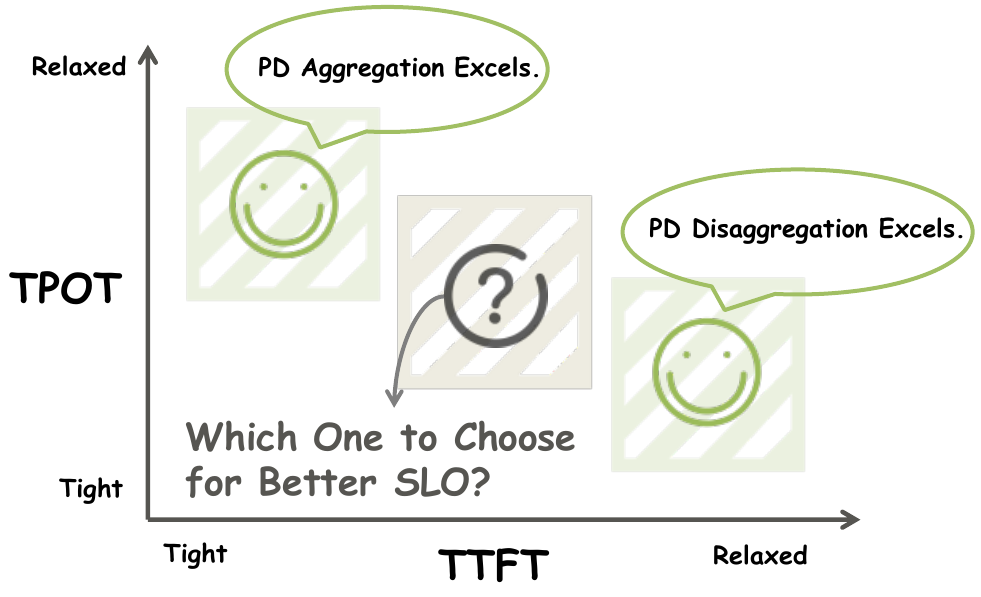}
    \caption{\revise{Trade-off between PD aggregation and disaggregation under different TTFT and TPOT constraints.}}
    \label{fig:pd_tradeoff}
\end{figure}

Large language model (LLM) serving has become a dominant workload in modern datacenters, and its cost is increasingly dominated by how efficiently the underlying GPUs are utilized~\cite{orca,vllm,sarathi}. A request in an LLM service is processed in two phases. The prefill phase reads the entire prompt in one compute-saturating pass and produces the first token, and its latency is bounded by the time-to-first-token (TTFT). The decode phase then emits output tokens one at a time while the KV cache grows in GPU memory, and its latency is bounded by the time-per-output-token (TPOT). Since prefill is compute-bound and decode is memory-bound, co-locating them on the same GPU causes mutual interference~\cite{sarathi,splitwise}.

\revise{This observation has motivated prefill-decode (PD) disaggregation, which places the two phases on separate worker pools and transfers the KV cache between them~\cite{distserve,splitwise,mooncake}. These systems address phase placement, resource allocation, and cache management. We study the complementary question of choosing local or remote prefill for each request under a given P/D allocation.}

\revise{TaiChi's simulated ArXiv workload at 12 requests/s quantifies the trade-off illustrated in Figure~\ref{fig:pd_tradeoff}~\cite{taichi}. With TTFT/TPOT targets of 5\,s/250\,ms, aggregation and disaggregation achieve SLO attainment rates of 97\% and 42\%, respectively. With targets of 16\,s/60\,ms, these rates become 7\% and 98\%. At 6\,s/100\,ms, they are 16\% and 50\%. Aggregation allows all instances to process prefill, but prefill and decode share compute resources. Disaggregation isolates the two phases while assigning only a subset of instances to prefill. These results motivate request-level path selection based on prefill cost, cache reuse, and decode demand.}

\revise{An arriving request can either run both phases on a decode worker or run prefill on a prefill worker followed by decode on a decode worker. NVIDIA Dynamo exposes this choice through configurable rules based on effective input length, the effective-to-raw input-length ratio, and prefill load~\cite{dynamo}. Our evaluated baseline uses the length-based rule described in Section~\ref{sec:baselines}. We investigate learning cluster-level routing boundaries from measured SLO feedback using cache information from both worker pools and output-length information.}

\revise{To address this, we introduce DolphinRouting, which learns cluster-level binary routing decisions from SLO feedback. We group requests with similar input lengths, output lengths, and cache-reuse patterns into $K$ clusters offline and learn one binary decision per cluster.} At runtime, an incoming request is assigned to its cluster, and the corresponding decision determines whether it follows the aggregated or disaggregated path. \revise{This path selection uses feature extraction, cluster assignment, and table lookup, avoiding an additional per-request cost-model scoring step. Instance-level scheduling and execution remain with the serving backend. We evaluate DolphinRouting with two prefill/decode allocations on the test cluster, learning a separate decision table for each allocation.}

\revise{We formulate the problem as cluster-level binary policy search, where a policy assigns one routing action to each of the $K$ clusters.} Each request is represented by four features: raw sequence length, P-side effective ISL, D-side effective ISL, and output sequence length. The decisions are learned in two stages, per-cluster scanning for initialization and $\epsilon$-greedy search for refinement. \revise{On Hybrid, DolphinRouting with $k$-means reduces P90 TTFT by 7.5--7.9\% and P90 TPOT by 6.5--10.1\% relative to Dynamo across the two allocations. On Real Data, the respective reductions are 5.7--6.5\% and 2.6--5.5\%.}

\section{Related Work}
\label{sec:related}

\subsection{Prefill-Decode Disaggregation}
\label{sec:related-pd}

Serving systems address prefill--decode interference through batching and adaptive execution. Orca introduces iteration-level scheduling and selective batching~\cite{orca}, and Sarathi-Serve uses chunked prefills and stall-free batching to limit interruptions to ongoing decodes~\cite{sarathi}. POD-Attention overlaps prefill and decode attention within a GPU~\cite{kamath2025pod}, while DuetServe combines aggregated execution with adaptive SM-level partitioning and activates spatial isolation when predicted contention threatens token-generation latency~\cite{gao2025duetserve}. These approaches schedule the two phases within shared GPU resources.

Disaggregated systems assign the phases to separate workers. DistServe co-optimizes phase placement, parallelism, and resource allocation for goodput~\cite{distserve}, and Splitwise uses phase-specific hardware on separate machines~\cite{splitwise}. TetriInfer combines fixed-size prompt chunks with two-level scheduling and predicted resource demands across decode instances~\cite{hu2024inference}. P/D-Serve scales disaggregated serving on Ascend clusters through flexible P/D organization, request forwarding, and device-to-device KV transfer~\cite{jin2024p}. These systems address resource provisioning, workload balance, and the cost of moving request state.

Other designs make the phase boundary more flexible. TaiChi unifies aggregation and disaggregation through differentiated-capability instances and a latency-shifting scheduler~\cite{taichi}. DynaServe splits requests into micro-requests according to the P/D resource ratio and instance load~\cite{dynaserve}, and Adrenaline offloads decode attention computation to prefill instances~\cite{adrenaline}. These approaches adjust where parts of a request execute.

Instance selection and migration operate at another scheduling level. Llumnix reschedules requests and migrates their in-memory state to balance serving instances~\cite{sun2024llumnix}. Calibrate, Then Route estimates additional completion time using prompt length, predicted output length, KV-cache pressure, and SLO class~\cite{routing}. These works determine which instance serves a request; DolphinRouting chooses whether prefill executes locally on a decode worker or remotely on a prefill worker.

Dynamo exposes this path choice through conditional disaggregation based on effective input length, prefill load, or their combination~\cite{dynamo}. Under a given P/D allocation, DolphinRouting groups requests using raw input length, cache-adjusted input lengths on both sides, and estimated output length. It learns one binary action per cluster from end-to-end SLO feedback and uses the resulting table online, leaving resource allocation and within-worker execution to the backend.

\subsection{KV Cache Transfer}
\label{sec:related-kv}

KV-cache management determines how much prefill computation can be reused and what state must move between workers. vLLM's PagedAttention manages and shares KV caches in blocks~\cite{vllm}, and SGLang's RadixAttention reuses KV caches across language-model calls~\cite{zheng2024sglang}. Raw prompt length therefore does not fully describe prefill work, since workers may hold different amounts of cached context for the same request. DolphinRouting captures this difference through separate P-side and D-side effective input lengths.

Reuse also extends across conversation turns and memory tiers. CachedAttention preserves conversation KV caches in the AttentionStore hierarchy and overlaps cache loading and saving with computation~\cite{gao2024cost}. MemServe combines context caching and disaggregated inference through an elastic memory pool and locality-aware scheduling~\cite{hu2024memserve}, and Mooncake coordinates cache reuse, transfer, and admission around a distributed KV-cache pool~\cite{mooncake_project}. These systems connect cache placement to scheduling, since exploiting remote or offloaded state requires accounting for retrieval and transfer.

CacheGen compresses KV tensors and adapts loading to available bandwidth~\cite{cachegen}. CacheBlend extends reuse to non-prefix text chunks in retrieval-augmented generation, combining cached chunks with selective recomputation~\cite{yao2025cacheblend}. Compression and selective recomputation change the communication and computation costs that a routing policy must weigh.

At the transport layer, NIXL provides a common interface across memory and storage backends~\cite{nixl}, and vLLM's NixlConnector supports P/D transfers, including bidirectional reuse that lets prefill workers retrieve state retained by decode workers from earlier turns~\cite{vllm-nixl}. FlowKV combines an optimized KV-transfer layout with load-aware scheduling and flexible P/D allocation~\cite{li2025flowkv}. Transfer optimization and routing are thus often coupled. DolphinRouting focuses on path selection from request and cache features and evaluates decisions through measured SLO attainment on the deployed system.

\subsection{Novelty of Our Work}
\label{sec:related-novelty}

DolphinRouting introduces a cache-aware, cluster-level policy for choosing the prefill execution path. DistServe optimizes phase placement and resource allocation~\cite{distserve}; our policy selects local or remote prefill under a given P/D allocation. Dynamo's conditional routing rules~\cite{dynamo} use effective input length or prefill load; DolphinRouting jointly considers raw prompt length, cache-adjusted lengths on both P and D workers, and output-length information, which distinguishes requests with different cache reuse and decode demands. Offline scans and joint-policy evaluations learn one binary action per request cluster from end-to-end SLO feedback. Runtime routing then uses feature extraction, cluster assignment, and table lookup. The design connects cache asymmetry to execution-path decisions while keeping policy search outside the request-serving path.

\section{Methodology}
\label{sec:methodology}

\begin{figure*}[!t]
    \centering
    \includegraphics[width=0.9\linewidth]{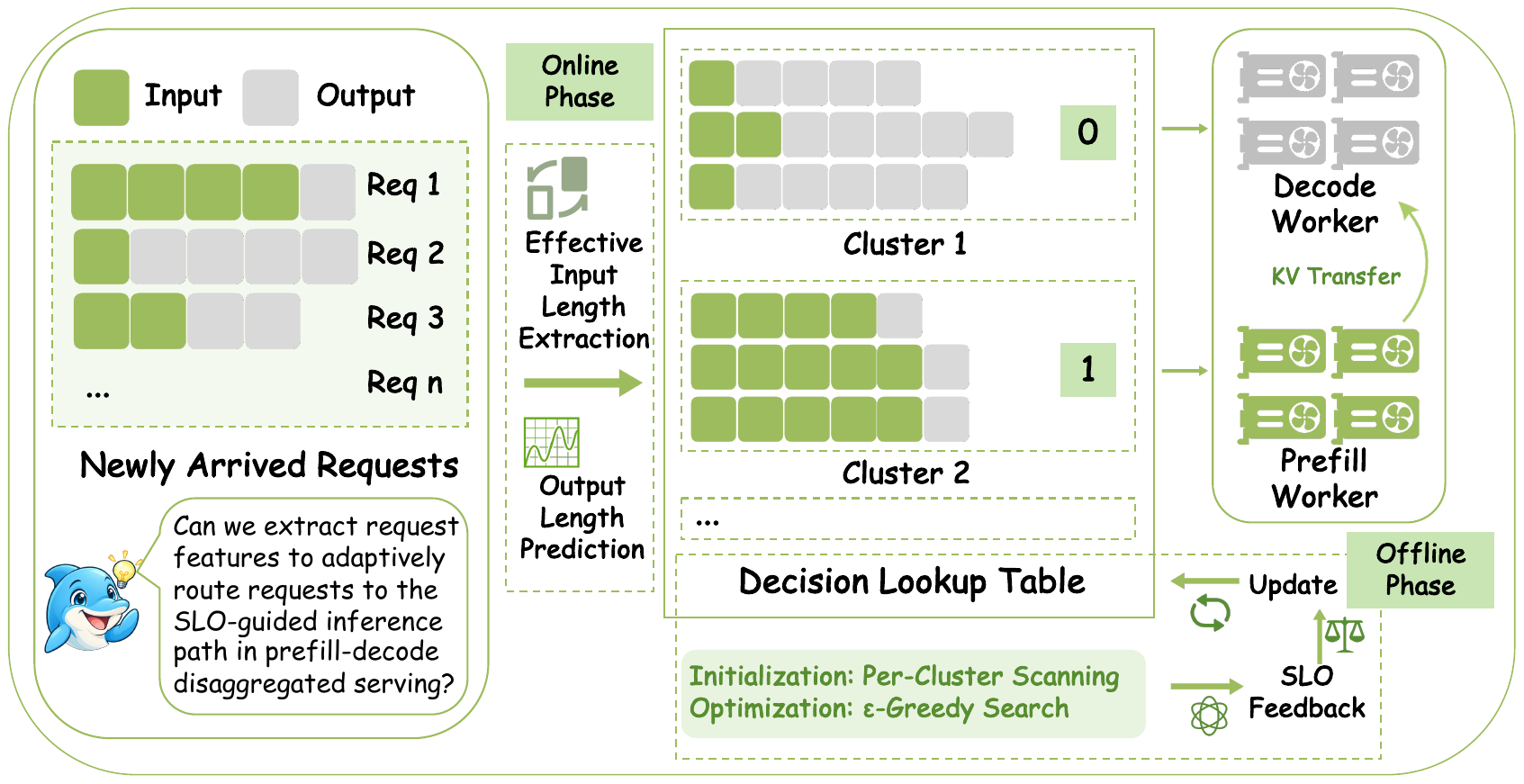}
    \caption{Overview of DolphinRouting.}
    \label{fig:main}
\end{figure*}

\subsection{Overview}
\label{sec:overview}

DolphinRouting learns cluster-level binary routing decisions for disaggregated LLM serving. The core idea is to partition incoming requests into $K$ behaviorally coherent clusters, and for each cluster learn a single binary decision: whether its requests should follow the \emph{aggregated} path (local prefill and decode on a decode worker) or the \emph{disaggregated} path (remote prefill on a prefill worker, followed by decode on a decode worker).

Formally, let $c \in \{1, \dots, K\}$ denote a cluster, and let
\begin{equation}
a_c \in \{0, 1\}, \quad
a_c =
\begin{cases}
0, & \text{aggregated path},\\
1, & \text{disaggregated path},
\end{cases}
\label{eq:action}
\end{equation}
denote the routing action assigned to cluster $c$. The \emph{cluster profile} summarizes the input-length distribution, output-length distribution, and cache-reuse pattern of the cluster's member requests. The reward of cluster $c$ is the SLO attainment rate of its requests.

A routing policy is a vector $\mathbf{a}=(a_1,\ldots,a_K)\in\{0,1\}^K$, giving $2^K$ possible decision tables. Because clusters share workers and caches, a cluster's reward depends on the joint policy, workload, and cache state. DolphinRouting uses per-cluster action estimates to guide a binary policy search, evaluating the resulting joint policies on the serving system.

\subsection{System Architecture}
\label{sec:architecture}

DolphinRouting consists of a feature extractor, a cluster assignment module, a routing decision lookup table, and a decision updater. At runtime, the extractor computes the raw prompt length, P-side and D-side effective ISL, and estimated output length. The assignment module maps this vector to a cluster ID, whose one-bit table entry selects the aggregated ($0$) or disaggregated ($1$) path.

All policy optimization is performed offline. Historical traces are used to fit the feature extractor and output-length predictor, group requests into $K$ clusters, and construct profiles of their length distributions and cache-reuse patterns. The decision updater initializes the table through per-cluster scanning and refines it with $\epsilon$-greedy search, using SLO feedback measured under saturated load.

\subsection{Feature Types}
\label{sec:features}

Four features capture prefill work, decode demand, and cache reuse.

\begin{itemize}
    \item \textbf{Request sequence length} $\ell^{\text{raw}}$: the raw prompt length before subtracting any cached prefix.
    \item \textbf{P-side effective ISL} $\ell^{P}$: the prompt length after subtracting the cached prefix overlap on the prefill worker, i.e., the prefill computation the prefill worker must perform on the disaggregated path.
    \item \textbf{D-side effective ISL} $\ell^{D}$: the prompt length after subtracting the cached prefix overlap on the decode worker, i.e., the local prefill computation the decode worker must perform on the aggregated path.
    \item \textbf{Output sequence length} $L$: the number of generated tokens, which reflects decode-side pressure and KV growth.
\end{itemize}

The two effective ISL features implicitly encode cache reuse, since they subtract the cache hit length from the raw prompt length. The output length captures output-side behavior that length features alone cannot reflect.

The offline stage uses ground-truth output lengths from historical traces. At admission, the output length is unknown, so we use a coarse estimate $\hat{L}$ from \emph{Clairvoyant}, a CPU-deployed XGBoost classifier that predicts short, medium, or long responses~\cite{sundaresan2026clairvoyant}. Clairvoyant uses 19 lexical features without a tokenizer, and its authors report $0.029$~ms per request, which is orders of magnitude smaller than LLM inference latency. We map the predicted classes to representative output lengths.

\subsection{Offline Clustering}
\label{sec:clustering}

Clustering determines which requests must share a routing action. We cluster requests using the four features with either $k$-means or a Gaussian mixture model (GMM), where $K$ is the number of clusters.

Algorithm~\ref{alg:clustering} summarizes the procedure. Feature extraction is parallelized across requests, since the raw sequence length, the two effective ISL values, and the output length come from independent sources. Offline, we use the ground-truth output length recorded in the historical traces instead of a prediction. The three length features are long-tailed and span several orders of magnitude, so we apply a $\log(1+x)$ transform followed by standardization before clustering. For each cluster, we build a profile that records the distributions of the four features and the cluster's request proportion in the historical workload.

Clustering groups requests with similar prefill cost and cache-reuse behavior, so that each cluster can receive a single binary decision that matches its dominant behavior. Requests with a small D-side effective ISL are cheap to serve locally on a decode worker, while routing them through the disaggregated path only adds transfer overhead. Requests with a large D-side effective ISL would otherwise impose a heavy prefill burden on the decode worker.

\begin{algorithm}[!tbp]
\caption{Offline Clustering and Profile Construction}
\label{alg:clustering}
\begin{algorithmic}[1]
\REQUIRE Historical traces $\mathcal{D} = \{x_1, \dots, x_N\}$
\ENSURE Cluster assignment $\phi$, cluster profiles $\{\mathcal{P}_c\}_{c=1}^{K}$
\STATE \textbf{Feature extraction (parallel):}
\FOR{$x_i \in \mathcal{D}$ \textbf{in parallel}}
    \STATE $\ell_i^{\text{raw}} \leftarrow \mathrm{len}(x_i)$
    \STATE $\ell_i^{P} \leftarrow \ell_i^{\text{raw}} - \mathrm{Hit}_{P}(x_i)$
    \STATE $\ell_i^{D} \leftarrow \ell_i^{\text{raw}} - \mathrm{Hit}_{D}(x_i)$
    \STATE $L_i \leftarrow \mathrm{OSL}(x_i)$
\ENDFOR
\STATE \textbf{Normalization:} $\ell^{(\cdot)} \leftarrow \mathrm{ZScore}\bigl(\log(1+\ell^{(\cdot)})\bigr)$ for length features
\STATE \textbf{Clustering:} $\phi \leftarrow \mathrm{Cluster}\bigl(\{x_i\}_{i=1}^{N}\bigr)$
\FOR{$c = 1, \dots, K$}
    \STATE $\mathcal{P}_c \leftarrow \mathrm{Profile}\bigl(\{x_i : \phi(x_i) = c\}\bigr)$
    \IF{$\mathrm{Span}(\mathcal{P}_c) > \tau$}
        \STATE $\mathrm{Mark}(c)$
    \ENDIF
\ENDFOR
\STATE \textbf{return} $\phi$, $\{\mathcal{P}_c\}_{c=1}^{K}$
\end{algorithmic}
\end{algorithm}

\subsection{Reward Design}
\label{sec:reward}

We optimize the fraction of requests that meet an end-to-end completion-time SLO. For request $r$, let $T_r^{\mathrm{e2e}}$ denote the elapsed time from request arrival to generation of the last output token, and let $\text{SLO}_r$ denote its end-to-end latency threshold. The reward for cluster $c$ is
\begin{equation}
R_c = \frac{1}{|\mathcal{R}_c|} \sum_{r \in \mathcal{R}_c}
\mathbb{I}\!\left[
T_r^{\mathrm{e2e}} \leq \text{SLO}_r
\right],
\label{eq:reward}
\end{equation}
where $\mathcal{R}_c$ is the set of requests routed to cluster $c$ during reward measurement. A completed request with $L_r \geq 2$ output tokens has $T_r^{\mathrm{e2e}} = \text{TTFT}_r + (L_r-1)\,\text{TPOT}_r$.

\subsection{Decision Search}
\label{sec:search}

DolphinRouting's decision search consists of two stages: an initialization stage that produces a coarse decision table, and an optimization stage that refines it. Both stages measure SLO attainment under saturated load.

\paragraph{Initialization.}
The initialization stage adopts \emph{per-cluster scanning}, as summarized in Algorithm~\ref{alg:initialization}. For each cluster $c$, the decisions of all other clusters are held fixed, and the requests of cluster $c$ are routed through the aggregated path for a period $T$, after which the reward $R_{\text{agg}}(c)$ is recorded. The same cluster is then routed through the disaggregated path for the same duration $T$, and the reward $R_{\text{disagg}}(c)$ is recorded. The action with the higher reward is selected as the initial decision for cluster $c$. The $2K$ scans produce the initial decision table and a measured reward for each cluster-action pair, which initialize the subsequent search. These rewards reflect the other clusters' routing choices and cache state during each scan.

\begin{algorithm}[!tbp]
\caption{Initialization by Per-Cluster Scanning}
\label{alg:initialization}
\begin{algorithmic}[1]
\REQUIRE Cluster assignment $\phi$, number of clusters $K$, scanning duration $T$, saturated load
\ENSURE Initial decision table $\{a_c\}_{c=1}^{K}$ and scan rewards $\{R_{\text{agg}}(c), R_{\text{disagg}}(c)\}_{c=1}^{K}$
\FOR{$c = 1, \dots, K$}
    \STATE Fix $\{a_{c'}\}_{c' \neq c}$
    \STATE Route cluster $c$ via the aggregated path for $T$ and record $R_{\text{agg}}(c)$
    \STATE Route cluster $c$ via the disaggregated path for $T$ and record $R_{\text{disagg}}(c)$
    \STATE $a_c \leftarrow \arg\max_{a \in \{0,1\}} \{R_{\text{agg}}(c), R_{\text{disagg}}(c)\}$
\ENDFOR
\STATE \textbf{return} $\{a_c\}_{c=1}^{K}$, $\{R_{\text{agg}}(c), R_{\text{disagg}}(c)\}_{c=1}^{K}$
\end{algorithmic}
\end{algorithm}

\paragraph{Optimization.}
The optimization stage adopts \emph{$\epsilon$-greedy per-cluster search}, as summarized in Algorithm~\ref{alg:epsilon}. The scan rewards initialize $\hat{R}_c(0)=R_{\text{agg}}(c)$ and $\hat{R}_c(1)=R_{\text{disagg}}(c)$. Since each action has been measured once, we initialize $n_c(0)=n_c(1)=1$, where $n_c(a)$ counts reward-measurement windows. The first greedy choice therefore uses the preferences established by scanning. For every cluster $c$, with probability $\epsilon_t$ the cluster selects a random action in $\{0,1\}$, and with probability $1-\epsilon_t$ it selects the action with the higher estimated reward $\hat{R}_c(a)$. The requests are then routed according to the resulting decision table under saturated load, and the SLO attainment is observed. The reward estimate of each cluster is updated using an incremental average over the action that was actually executed.

The exploration probability follows an exponential decay schedule,
\begin{equation}
\epsilon_t = \epsilon_{\min} + (\epsilon_0 - \epsilon_{\min}) \exp\!\left(-\frac{t}{K}\right),
\label{eq:epsilon}
\end{equation}
with $\epsilon_0 = 0.5$ and $\epsilon_{\min} = 0.01$. Using $K$ as the decay time constant keeps exploration active for more rounds as the number of binary decisions grows.

\begin{algorithm}[!tbp]
\caption{Optimization by $\epsilon$-Greedy Search}
\label{alg:epsilon}
\begin{algorithmic}[1]
\REQUIRE Initial decision table $\{a_c\}_{c=1}^{K}$, scan rewards $\{R_{\text{agg}}(c), R_{\text{disagg}}(c)\}_{c=1}^{K}$, horizon $T_{\max}$, saturated load
\ENSURE Final decision table $\{a_c\}_{c=1}^{K}$
\STATE $\hat{R}_c(0) \leftarrow R_{\text{agg}}(c)$ for all $c$
\STATE $\hat{R}_c(1) \leftarrow R_{\text{disagg}}(c)$ for all $c$
\STATE $n_c(0) \leftarrow 1$, $n_c(1) \leftarrow 1$ for all $c$
\FOR{$t = 1, \dots, T_{\max}$}
    \STATE $\epsilon_t \leftarrow \epsilon_{\min} + (\epsilon_0 - \epsilon_{\min}) \exp(-t/K)$
    \FOR{$c = 1, \dots, K$ \textbf{in parallel}}
        \STATE $a_c \leftarrow$ random action in $\{0,1\}$ with probability $\epsilon_t$; otherwise $a_c \leftarrow \arg\max_{a} \hat{R}_c(a)$
    \ENDFOR
    \STATE Route requests according to $\{a_c\}_{c=1}^{K}$ under saturated load; observe SLO attainment
    \FOR{$c = 1, \dots, K$}
        \STATE $R_c \leftarrow$ cluster reward for the current round
        \STATE $n_c(a_c) \leftarrow n_c(a_c) + 1$
        \STATE $\hat{R}_c(a_c) \leftarrow \hat{R}_c(a_c) + \frac{1}{n_c(a_c)}\bigl(R_c - \hat{R}_c(a_c)\bigr)$
    \ENDFOR
\ENDFOR
\STATE \textbf{return} $\{a_c\}_{c=1}^{K}$
\end{algorithmic}
\end{algorithm}
\section{Experiments}
\label{sec:experiments}

\subsection{Testbed}
\label{sec:testbed}

All experiments use \texttt{Qwen3.8-72B} \cite{qwen38} as the served model and are conducted on a cluster where each node is equipped with 4 NVIDIA H20-SXM GPUs, each with 141\,GB HBM3e, connected via NVLink with 900\,GB/s interconnect bandwidth. Each node also carries an Intel Xeon Platinum 8457C CPU and RDMA network cards supporting up to 800\,Gb/s of inter-node bandwidth. Each node deploys either a prefill instance or a decode instance according to the startup parameter.

\subsection{Datasets and Workloads}
\label{sec:datasets}

We select three public benchmarks and one real-world trace as outlined in Table~\ref{tab:datasets}. ArXiv Summarization \cite{arxiv-sum} represents a prefill-heavy workload with long inputs and short outputs. L-Eval \cite{leval} represents a decode-reuse-heavy workload with very long inputs but high prefix-cache hit rates, so that the effective prefill cost is much smaller than the raw input length suggests. HumanEval \cite{humaneval} represents a short-input, short-output, cache-cold workload, where disaggregation offers little benefit. The Hybrid workload is constructed by mixing the three benchmarks in proportion to their Poisson arrival intensities, so that it contains both prefill-heavy and decode-heavy request types. The Real Data workload is drawn from the publicly available conversation trace in Mooncake~\cite{mooncake_project}, which contains authentic user requests with realistic input/output length distributions.

\begin{table}[htbp]
\centering
\small
\caption{Dataset characteristics. Lengths are average token counts.}
\label{tab:datasets}
\begin{tabular}{@{}lcccl@{}}
\toprule
Dataset & \shortstack{Input\\length} & \shortstack{Output\\length} & \shortstack{Cache\\ratio} & Arrival \\
\midrule
ArXiv Summarization & 8088 & 229 & $\sim$0\% & Poisson \\
L-Eval & 19019 & 72 & $>$80\% & Poisson \\
HumanEval & 198 & 221 & $\sim$0\% & Poisson \\
Real Data & 12035 & 343 & $\sim$50\% & Timestamp \\
\bottomrule
\end{tabular}
\end{table}

\begin{table*}[!t]
\centering
\small
\caption{Normalized P90 TTFT and TPOT at the saturated operating point under five workloads and two PD settings.}
\label{tab:main}
\begin{tabular*}{\textwidth}{@{\extracolsep{\fill}}llcccc@{}}
\toprule
\multirow{2}{*}{Workload} & \multirow{2}{*}{Method} &
\multicolumn{2}{c}{P3D1} & \multicolumn{2}{c}{P2D2} \\
\cmidrule(lr){3-4} \cmidrule(lr){5-6}
& & Norm TTFT P90 $\downarrow$ & Norm TPOT P90 $\downarrow$ & Norm TTFT P90 $\downarrow$ & Norm TPOT P90 $\downarrow$ \\
\midrule
\multirow{4}{*}{ArXiv}
& Naive          & 1.00 & 0.49 & 1.00 & 0.22 \\
& Dynamo         & 0.99 & 0.50 & 1.01 & 0.24 \\
& DolphinRouting ($k$-means) & 0.98 & 0.51 & 1.03 & 0.24 \\
& DolphinRouting (GMM)     & 0.99 & 0.50 & 1.02 & 0.23 \\
\midrule
\multirow{4}{*}{L-Eval}
& Naive          & 0.21 & 1.00 & 0.99 & 0.25 \\
& Dynamo         & 0.24 & 0.92 & 0.91 & 0.27 \\
& DolphinRouting ($k$-means) & 0.25 & 0.90 & 0.85 & 0.28 \\
& DolphinRouting (GMM)     & 0.24 & 0.92 & 0.90 & 0.27 \\
\midrule
\multirow{4}{*}{HumanEval}
& Naive          & 0.12 & 1.01 & 0.19 & 1.02 \\
& Dynamo         & 0.11 & 1.03 & 0.19 & 1.03 \\
& DolphinRouting ($k$-means) & 0.13 & 1.04 & 0.20 & 1.06 \\
& DolphinRouting (GMM)     & 0.13 & 1.03 & 0.21 & 1.05 \\
\midrule
\multirow{4}{*}{Hybrid}
& Naive          & 0.42 & 1.00 & 1.01 & 0.30 \\
& Dynamo         & 0.40 & 0.89 & 0.89 & 0.31 \\
& DolphinRouting ($k$-means) & 0.37 & \underline{0.80} & \underline{0.82} & 0.29 \\
& DolphinRouting (GMM)     & 0.39 & 0.83 & 0.84 & 0.30 \\
\midrule
\multirow{4}{*}{Real Data}
& Naive          & 0.38 & 1.00 & 1.02 & 0.42 \\
& Dynamo         & 0.35 & 0.91 & 0.92 & 0.39 \\
& DolphinRouting ($k$-means) & 0.33 & \underline{0.86} & \underline{0.86} & 0.38 \\
& DolphinRouting (GMM)     & 0.35 & 0.89 & 0.87 & 0.37 \\
\bottomrule
\end{tabular*}
\end{table*}

% Declare the wide latency plots early enough to accompany the results.
\begin{figure*}[!t]
\centering
\begin{minipage}[t]{0.48\linewidth}
    \centering
    \includegraphics[width=\linewidth]{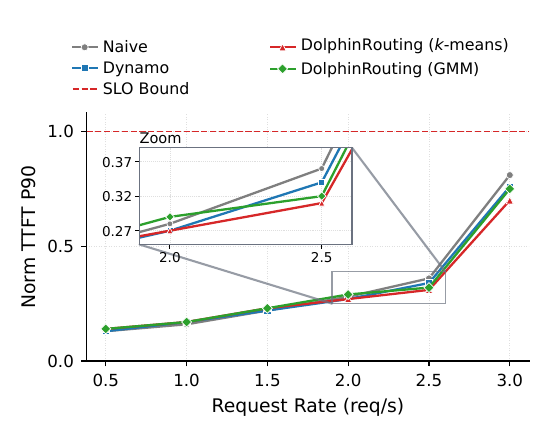}
\end{minipage}
\hfill
\begin{minipage}[t]{0.48\linewidth}
    \centering
    \includegraphics[width=\linewidth]{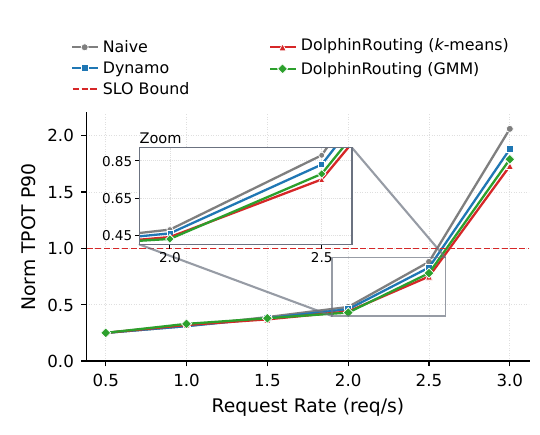}
\end{minipage}
\caption{Normalized P90 TTFT and P90 TPOT on the Hybrid workload under P3D1 as the request rate increases. The dashed lines at 1.0 mark the SLO bounds. Insets magnify the boxed regions.}
\label{fig:latency-p3d1}
\end{figure*}

\begin{figure*}[!t]
\centering
\begin{minipage}[t]{0.48\linewidth}
    \centering
    \includegraphics[width=\linewidth]{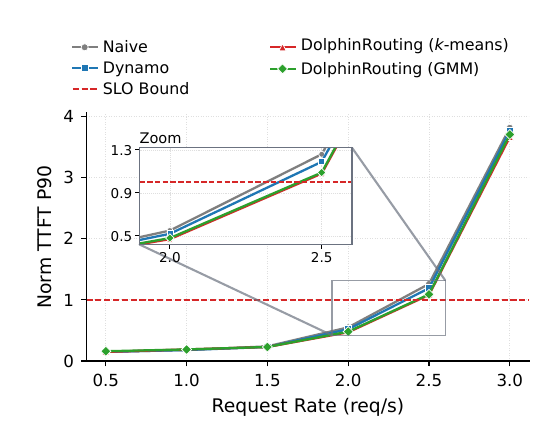}
\end{minipage}
\hfill
\begin{minipage}[t]{0.48\linewidth}
    \centering
    \includegraphics[width=\linewidth]{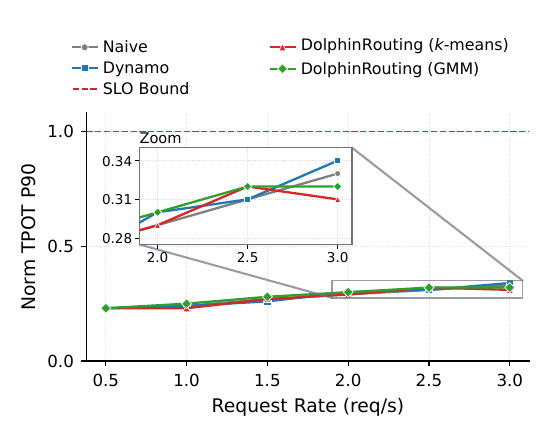}
\end{minipage}
\caption{Normalized P90 TTFT and P90 TPOT on the Hybrid workload under P2D2 as the request rate increases. The dashed lines at 1.0 mark the SLO bounds. Insets magnify the boxed regions.}
\label{fig:latency-p2d2}
\end{figure*}

\subsection{Metrics}
\label{sec:metrics}

We report two metrics, both measured at the saturated operating point described below. \textbf{TTFT (Time To First Token)} is the time from request arrival to the generation of the first output token, which reflects the combined cost of prefill computation, queueing, and KV transfer. \textbf{TPOT (Time Per Output Token)} is the average time between consecutive output tokens during the decode stage, which reflects the decode-side load and KV-cache pressure. We use the 90th percentile (P90) of both metrics, which captures whether the majority of requests satisfy the SLOs and is more representative of the user-perceived tail behavior.

For each workload and PD configuration, we measure the Naive configuration at a near-idle request rate of $0.1$ requests/s, and use its P90 TTFT and P90 TPOT as the reference latencies $b_{\mathrm{TTFT}}$ and $b_{\mathrm{TPOT}}$, respectively. The SLO bounds are $B_{\mathrm{TTFT}} = 10b_{\mathrm{TTFT}}$ and $B_{\mathrm{TPOT}} = 5b_{\mathrm{TPOT}}$, and are fixed across methods within the same workload and PD configuration. Exceeding these bounds indicates a failure to meet the SLO, and the corresponding consumed resources are considered wasted. We normalize the measured TTFT and TPOT values against these bounds, so that a normalized value of 1.0 corresponds to the SLO boundary, values below 1.0 indicate that the SLO is satisfied, and lower values are better.

All reported results correspond to the saturated operating point for each workload. Specifically, we increase the request rate from a low value and stop at the lower boundary of saturation, i.e., the smallest load at which $\mathrm{P90}(\mathrm{TTFT}) \ge B_{\mathrm{TTFT}}$ or $\mathrm{P90}(\mathrm{TPOT}) \ge B_{\mathrm{TPOT}}$, equivalently $\mathrm{Norm\ P90\ TTFT} \ge 1.0$ or $\mathrm{Norm\ P90\ TPOT} \ge 1.0$. The main table reports the metrics at this operating point.

\subsection{Baselines}
\label{sec:baselines}

All systems are instantiated on vLLM as the serving backend, and use Mooncake~\cite{mooncake} as the KV transfer mechanism for the disaggregated path. We compare four systems. \textbf{Naive} is the original PD-disaggregated configuration without conditional routing, and serves as the default baseline. \textbf{Dynamo} denotes the length-based conditional disaggregation rule evaluated here~\cite{dynamo}: a request takes the aggregated path when its effective ISL is below a global threshold and its effective-to-raw ISL ratio is below a global ratio threshold. This comparison does not cover Dynamo's additional load-based policies. \textbf{DolphinRouting ($k$-means)} and \textbf{DolphinRouting (GMM)} instantiate DolphinRouting with $k$-means and Gaussian mixture model clustering, respectively.

\subsection{Settings}
\label{sec:settings}

We evaluate each system under two PD settings, namely \textbf{P3D1} (three prefill instances and one decode instance) and \textbf{P2D2} (two prefill instances and two decode instances). For each workload and PD configuration, we first determine the saturated request rate on the Naive configuration, and then compare all methods at this same request rate. For DolphinRouting, both clustering and decision optimization are performed offline on a 20\% sample of each dataset. This sample is drawn from a separate trace that does not overlap with the requests used for evaluation. We set the number of clusters to $K = 8$ in all experiments. The sensitivity to $K$ is examined in Section~\ref{sec:ablation}. Each configuration is repeated three times, and the reported numbers are the average across the three runs.

\subsection{Main Results}
\label{sec:main}

Table~\ref{tab:main} compares normalized P90 TTFT and TPOT at the saturated operating point, with all methods evaluated at the same operating point within each workload and PD configuration. Values are normalized by the SLO bounds defined in Section~\ref{sec:metrics}, so 1.0 marks the SLO boundary and lower is better.

On ArXiv, DolphinRouting is close to Naive. With $k$-means, TTFT changes by about 2--3\% and TPOT by about 4--9\% across the two PD settings, and GMM shows similar small changes. On L-Eval, the two settings behave differently: under P3D1, $k$-means trades a higher TTFT for a lower TPOT, while under P2D2 it lowers TTFT and raises TPOT. These two workloads are dominated by a single request type, and routing shifts load between the prefill and decode sides without a consistent gain.

The benefit appears on the mixed and real traces. On Hybrid and Real Data, DolphinRouting ($k$-means) improves both metrics over Dynamo in all four settings, with TTFT reductions of roughly 6--8\% and TPOT reductions of roughly 3--10\%. Relative to Naive, the TTFT reduction reaches about 12--19\% and the TPOT reduction about 3--20\%. Neither metric improves over Dynamo in any of the four ArXiv/L-Eval settings, which is consistent with the lack of mixed prefill/decode demand in those workloads.

Between the two clustering choices, $k$-means is better in seven of the eight metric/configuration comparisons on Hybrid and Real Data; the exception is Real Data TPOT under P2D2, where GMM is marginally better (0.37 versus 0.38). The ablations below therefore use $k$-means.

\subsection{Analysis}
\label{sec:analysis}

Figures~\ref{fig:latency-p3d1} and~\ref{fig:latency-p2d2} report normalized P90 latencies on Hybrid as the request rate increases. Under P3D1, at 2.5 requests/s, $k$-means gives a TPOT of 0.75, below Naive (0.88) and Dynamo (0.83). When the rate rises to 3.0 requests/s, its TPOT crosses the SLO bound and reaches 1.73, and TTFT rises from 0.31 to 0.70. Under P2D2, increasing the rate from 2.0 to 2.5 requests/s pushes $k$-means TTFT from 0.47 to 1.08; at 2.5 requests/s this is below both Naive (1.26) and Dynamo (1.19), while TPOT stays near 0.32 and remains well below the SLO bound throughout the sweep.

The routing decision changes where prefill is executed. Local prefill uses decode-worker compute and avoids the remote-prefill KV transfer, while the disaggregated path executes prefill on a prefill worker and still uses a decode worker for generation. Learning a separate decision table for each allocation lets the policy choose between these costs using measured rewards.

\Needspace{6\baselineskip}
\subsection{Ablation Study}
\label{sec:ablation}

\subsubsection{Effect of the Number of Clusters}

The number of clusters $K$ controls the granularity of routing and the number of binary decisions to be learned. On the Hybrid workload, we vary $K \in \{2, 4, 8, 16\}$ and report the average number of optimization rounds needed to converge and the resulting normalized metrics under P3D1 and P2D2.

\begin{figure*}[t]
\centering
\begin{minipage}[t]{0.48\linewidth}
    \centering
    \includegraphics[width=\linewidth]{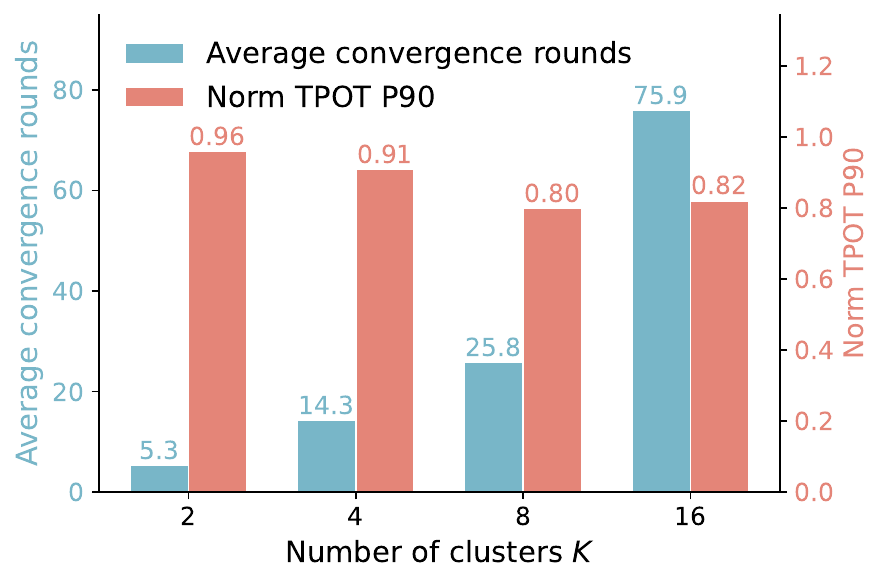}
\end{minipage}
\hfill
\begin{minipage}[t]{0.48\linewidth}
    \centering
    \includegraphics[width=\linewidth]{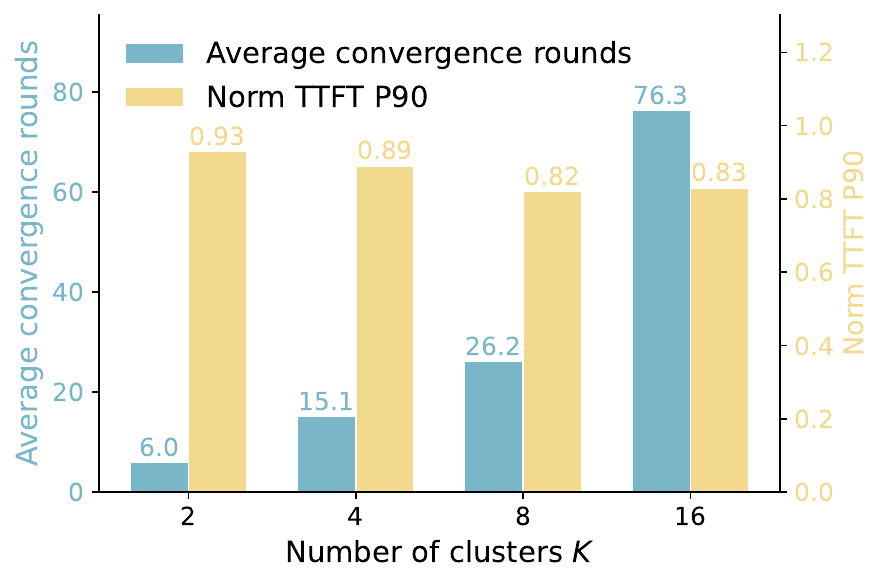}
\end{minipage}
\caption{Effect of the number of clusters $K$ on the average convergence rounds and the normalized P90 metric, under P3D1 (left) and P2D2 (right).}
\label{fig:ablation-k}
\end{figure*}

As shown in Figure~\ref{fig:ablation-k}, increasing $K$ from 8 to 16 roughly triples the optimization rounds, from about 26 to about 76 under both settings, while the normalized metrics change only slightly (TPOT 0.80 to 0.82 under P3D1; TTFT 0.82 to 0.83 under P2D2). Reducing $K$ to 2 or 4 shortens convergence to roughly 5--15 rounds but raises the corresponding latency metrics by about 9--20\% relative to $K=8$. We use $K=8$ in the remaining experiments.

\subsubsection{Effect of the Initialization Stage}

To isolate the contribution of initialization, we fix $K=8$ under P3D1 on Hybrid. We compare the full pipeline, which uses scanning followed by $\epsilon$-greedy search, with a variant that omits scanning and starts optimization from a random decision table.

\begin{table}[htbp]
\centering
\caption{Effect of the initialization stage under P3D1.}
\label{tab:ablation-init}
\begin{tabular}{lcc}
\toprule
Metric & Full pipeline & w/o initialization \\
\midrule
Avg conv. rounds $\downarrow$ & 25.8 & 39.6 \\
Norm TTFT P90 $\downarrow$ & 0.37 & 0.40 \\
Norm TPOT P90 $\downarrow$ & 0.80 & 0.91 \\
\bottomrule
\end{tabular}
\end{table}

Removing initialization increases the average number of optimization rounds from 25.8 to 39.6 (Table~\ref{tab:ablation-init}), and raises normalized TTFT from 0.37 to 0.40 and TPOT from 0.80 to 0.91. At $K=8$, initialization uses 16 scans to seed both action estimates per cluster, which explains most of the convergence gap.

\subsubsection{Effect of the Output-Length Feature}
We evaluate the contribution of the output-length feature by removing it from the clustering representation. With $K=8$ on Hybrid, the resulting variant uses three features: raw sequence length, P-side effective ISL, and D-side effective ISL.

\begin{table}[htbp]
\centering
\caption{Effect of the output-length (OSL) feature.}
\label{tab:ablation-osl}
\begin{tabular}{lcc}
\toprule
Metric & Full pipeline & w/o OSL \\
\midrule
\multicolumn{3}{l}{\textit{P3D1 setting}} \\
\midrule
Norm TTFT P90 $\downarrow$ & 0.37 & 0.42 \\
Norm TPOT P90 $\downarrow$ & 0.80 & 0.93 \\
\midrule
\multicolumn{3}{l}{\textit{P2D2 setting}} \\
\midrule
Norm TTFT P90 $\downarrow$ & 0.82 & 0.92 \\
Norm TPOT P90 $\downarrow$ & 0.29 & 0.30 \\
\bottomrule
\end{tabular}
\end{table}

Removing the output-length feature degrades both metrics in both settings (Table~\ref{tab:ablation-osl}). The effect is largest under P3D1, where normalized TTFT rises from 0.37 to 0.42 and TPOT from 0.80 to 0.93; under P2D2 the TTFT increase is similar (0.82 to 0.92), while TPOT changes only marginally (0.29 to 0.30).

\Needspace{5\baselineskip}
\section{Conclusion}
\label{sec:conclusion}

We presented DolphinRouting, a cluster-level binary routing framework for prefill-decode disaggregated LLM serving. DolphinRouting groups requests by their execution characteristics, including raw prompt length, cache-adjusted input lengths on both worker pools, and output-length information, and learns one binary routing action per cluster from end-to-end SLO feedback. Offline policy search constructs the decision table, and runtime feature extraction, cluster assignment, and table lookup select local or remote prefill. On the mixed workload and real data under both tested P/D allocations, DolphinRouting with $k$-means reduces P90 TTFT by 5.7--7.9\% and P90 TPOT by 2.6--10.1\% relative to Dynamo. These results show that a compact, cache-aware routing table learned from SLO feedback can improve tail latency without changing the underlying P/D resource allocation.

% Keep all manuscript figures ahead of the bibliography.
\FloatBarrier
\bibliography{ref}

@article{sundaresan2026clairvoyant,
  title={Clairvoyant: Predictive SJF Scheduling to Mitigate Head-of-Line Blocking in Serial LLM Backends},
  author={Sundaresan, Aravind},
  journal={arXiv preprint arXiv:2606.07248},
  year={2026}
}

@inproceedings{arxiv-sum,
  title = "A Discourse-Aware Attention Model for Abstractive Summarization of Long Documents",
  author = "Cohan, Arman  and
    Dernoncourt, Franck  and
    Kim, Doo Soon  and
    Bui, Trung  and
    Kim, Seokhwan  and
    Chang, Walter  and
    Goharian, Nazli",
  booktitle = "Proceedings of the 2018 Conference of the North {A}merican Chapter of the Association for Computational Linguistics: Human Language Technologies, Volume 2 (Short Papers)",
  month = jun,
  year = "2018",
  address = "New Orleans, Louisiana",
  publisher = "Association for Computational Linguistics",
  url = "https://aclanthology.org/N18-2097",
  doi = "10.18653/v1/N18-2097",
  pages = "615--621",
}

@misc{leval,
      title={L-Eval: Instituting Standardized Evaluation for Long Context Language Models}, 
      author={Chenxin An and Shansan Gong and Ming Zhong and Mukai Li and Jun Zhang and Lingpeng Kong and Xipeng Qiu},
      year={2023},
      eprint={2307.11088},
      archivePrefix={arXiv},
      primaryClass={cs.CL}
}

@article{humaneval,
  title={Evaluating Large Language Models Trained on Code},
  author={Mark Chen and Jerry Tworek and Heewoo Jun and Qiming Yuan and Henrique Ponde de Oliveira Pinto and Jared Kaplan and Harri Edwards and Yuri Burda and Nicholas Joseph and Greg Brockman and Alex Ray and Raul Puri and Gretchen Krueger and Michael Petrov and Heidy Khlaaf and Girish Sastry and Pamela Mishkin and Brooke Chan and Scott Gray and Nick Ryder and Mikhail Pavlov and Alethea Power and Lukasz Kaiser and Mohammad Bavarian and Clemens Winter and Philippe Tillet and Felipe Petroski Such and Dave Cummings and Matthias Plappert and Fotios Chantzis and Elizabeth Barnes and Ariel Herbert-Voss and William Hebgen Guss and Alex Nichol and Alex Paino and Nikolas Tezak and Jie Tang and Igor Babuschkin and Suchir Balaji and Shantanu Jain and William Saunders and Christopher Hesse and Andrew N. Carr and Jan Leike and Josh Achiam and Vedant Misra and Evan Morikawa and Alec Radford and Matthew Knight and Miles Brundage and Mira Murati and Katie Mayer and Peter Welinder and Bob McGrew and Dario Amodei and Sam McCandlish and Ilya Sutskever and Wojciech Zaremba},
  year={2021},
  eprint={2107.03374},
  archivePrefix={arXiv},
  primaryClass={cs.LG}
}

@misc{mooncake,
      title={Mooncake: A KVCache-centric Disaggregated Architecture for LLM Serving}, 
      author={Ruoyu Qin and Zheming Li and Weiran He and Mingxing Zhang and Yongwei Wu and Weimin Zheng and Xinran Xu},
      year={2026},
      eprint={2407.00079v5},
      archivePrefix={arXiv},
      primaryClass={cs.DC},
      url={https://arxiv.org/abs/2407.00079v5},
}

@inproceedings{mooncake_project,
  author    = {Ruoyu Qin and Zheming Li and Weiran He and Jialei Cui and Feng Ren and Mingxing Zhang and Yongwei Wu and Weimin Zheng and Xinran Xu},
  title     = {Mooncake: Trading More Storage for Less Computation {\textemdash} A {KVCache-centric} Architecture for Serving {LLM} Chatbot},
  booktitle = {23rd USENIX Conference on File and Storage Technologies (FAST 25)},
  year      = {2025},
  isbn      = {978-1-939133-45-8},
  address   = {Santa Clara, CA},
  pages     = {155--170},
  url       = {https://www.usenix.org/conference/fast25/presentation/qin},
  publisher = {USENIX Association},
  month     = {feb},
}

@misc{dynamo,
  title        = {NVIDIA Dynamo: An Open Source Inference Framework for Serving Generative AI},
  author       = {{NVIDIA}},
  year         = {2026},
  howpublished = {\url{https://docs.nvidia.com/dynamo/}},
  note         = {Accessed: 2026-10-26}
}

@misc{vllm,
      title={Efficient Memory Management for Large Language Model Serving with PagedAttention}, 
      author={Woosuk Kwon and Zhuohan Li and Siyuan Zhuang and Ying Sheng and Lianmin Zheng and Cody Hao Yu and Joseph E. Gonzalez and Hao Zhang and Ion Stoica},
      year={2023},
      eprint={2309.06180},
      archivePrefix={arXiv},
      primaryClass={cs.LG},
      url={https://arxiv.org/abs/2309.06180}, 
}

@misc{sarathi,
      title={Taming Throughput-Latency Tradeoff in LLM Inference with Sarathi-Serve}, 
      author={Amey Agrawal and Nitin Kedia and Ashish Panwar and Jayashree Mohan and Nipun Kwatra and Bhargav S. Gulavani and Alexey Tumanov and Ramachandran Ramjee},
      year={2024},
      eprint={2403.02310},
      archivePrefix={arXiv},
      primaryClass={cs.LG},
      url={https://arxiv.org/abs/2403.02310}, 
}

@misc{splitwise,
      title={Splitwise: Efficient generative LLM inference using phase splitting}, 
      author={Pratyush Patel and Esha Choukse and Chaojie Zhang and Aashaka Shah and Íñigo Goiri and Saeed Maleki and Ricardo Bianchini},
      year={2024},
      eprint={2311.18677},
      archivePrefix={arXiv},
      primaryClass={cs.AR},
      url={https://arxiv.org/abs/2311.18677}, 
}

@misc{distserve,
      title={DistServe: Disaggregating Prefill and Decoding for Goodput-optimized Large Language Model Serving}, 
      author={Yinmin Zhong and Shengyu Liu and Junda Chen and Jianbo Hu and Yibo Zhu and Xuanzhe Liu and Xin Jin and Hao Zhang},
      year={2024},
      eprint={2401.09670},
      archivePrefix={arXiv},
      primaryClass={cs.DC},
      url={https://arxiv.org/abs/2401.09670}, 
}

@misc{taichi,
      title={Prefill-Decode Aggregation or Disaggregation? Unifying Both for Goodput-Optimized LLM Serving}, 
      author={Chao Wang and Pengfei Zuo and Zhangyu Chen and Yunkai Liang and Zhou Yu and Ming-Chang Yang},
      year={2025},
      eprint={2508.01989},
      archivePrefix={arXiv},
      primaryClass={cs.DC},
      url={https://arxiv.org/abs/2508.01989}, 
}

@misc{routing,
      title={Calibrate, Then Route: A Measured Study of Learned Request Routing for Disaggregated LLM Serving}, 
      author={Srikanta Datta Tumkur and Jay Iyer and Mehar Simhadri and Sai Pavan Kumar and Sai Kapil Kumar and Ramesh Nampelly},
      year={2026},
      eprint={2609.16206},
      archivePrefix={arXiv},
      primaryClass={cs.AI},
      url={https://arxiv.org/abs/2609.16206}, 
}

@inproceedings {orca,
author = {Gyeong-In Yu and Joo Seong Jeong and Geon-Woo Kim and Soojeong Kim and Byung-Gon Chun},
title = {Orca: A Distributed Serving System for {Transformer-Based} Generative Models},
booktitle = {16th USENIX Symposium on Operating Systems Design and Implementation (OSDI 22)},
year = {2022},
isbn = {978-1-939133-28-1},
address = {Carlsbad, CA},
pages = {521--538},
url = {https://www.usenix.org/conference/osdi22/presentation/yu},
publisher = {USENIX Association},
month = jul
}

@misc{adrenaline,
      title={Injecting Adrenaline into LLM Serving: Boosting Resource Utilization and Throughput via Attention Disaggregation}, 
      author={Yunkai Liang and Zhangyu Chen and Pengfei Zuo and Zhi Zhou and Xu Chen and Zhou Yu},
      year={2025},
      eprint={2503.20552},
      archivePrefix={arXiv},
      primaryClass={cs.DC},
      url={https://arxiv.org/abs/2503.20552}, 
}

@misc{dynaserve,
      title={DynaServe: Unified and Elastic Execution for Dynamic Disaggregated LLM Serving}, 
      author={Chaoyi Ruan and Yinhe Chen and Dongqi Tian and Yandong Shi and Yongji Wu and Jialin Li and Cheng Li},
      year={2025},
      eprint={2504.09285},
      archivePrefix={arXiv},
      primaryClass={cs.DC},
      url={https://arxiv.org/abs/2504.09285}, 
}

@misc{cachegen,
      title={CacheGen: KV Cache Compression and Streaming for Fast Large Language Model Serving}, 
      author={Yuhan Liu and Hanchen Li and Yihua Cheng and Siddhant Ray and Yuyang Huang and Qizheng Zhang and Kuntai Du and Jiayi Yao and Shan Lu and Ganesh Ananthanarayanan and Michael Maire and Henry Hoffmann and Ari Holtzman and Junchen Jiang},
      year={2024},
      eprint={2310.07240},
      archivePrefix={arXiv},
      primaryClass={cs.NI},
      url={https://arxiv.org/abs/2310.07240}, 
}

@misc{nixl,
  title        = {{NVIDIA Inference Xfer Library (NIXL)}},
  author       = {{NVIDIA}},
  year         = {2025},
  howpublished = {\url{https://github.com/ai-dynamo/nixl}},
  note         = {Accessed: 2026-10-26}
}

@misc{vllm-nixl,
  title        = {{vLLM NixlConnector Usage Guide}},
  author       = {{vLLM Project}},
  year         = {2025},
  howpublished = {\url{https://docs.vllm.ai/en/latest/features/nixl_connector_usage/}},
  note         = {Accessed: 2026-10-26}
}

@misc{qwen38,
  title  = {{Qwen3.8-Max}: A New Bar for Coding and Cowork},
  author = {{Qwen Team}},
  year   = {2026},
  month  = aug,
  url    = {https://qwen.ai/blog?id=qwen3.8}
}

@inproceedings{sun2024llumnix,
  title={{Llumnix}: Dynamic Scheduling for Large Language Model Serving},
  author={Sun, Biao and Huang, Ziming and Zhao, Hanyu and Xiao, Wencong and Zhang, Xinyi and Li, Yong and Lin, Wei},
  booktitle={18th USENIX Symposium on Operating Systems Design and Implementation (OSDI 24)},
  pages={173--191},
  year={2024},
  url={https://www.usenix.org/conference/osdi24/presentation/sun-biao}
}

@article{hu2024inference,
  title={Inference without Interference: Disaggregate {LLM} Inference for Mixed Downstream Workloads},
  author={Hu, Cunchen and Huang, Heyang and Xu, Liangliang and Chen, Xusheng and Xu, Jiang and Chen, Shuang and Feng, Hao and Wang, Chenxi and Wang, Sa and Bao, Yungang and Sun, Ninghui and Shan, Yizhou},
  journal={arXiv preprint arXiv:2401.11181},
  year={2024},
  url={https://arxiv.org/abs/2401.11181}
}

@article{hu2024memserve,
  title={{MemServe}: Context Caching for Disaggregated {LLM} Serving with Elastic Memory Pool},
  author={Hu, Cunchen and Huang, Heyang and Hu, Junhao and Xu, Jiang and Chen, Xusheng and Xie, Tao and Wang, Chenxi and Wang, Sa and Bao, Yungang and Sun, Ninghui and Shan, Yizhou},
  journal={arXiv preprint arXiv:2406.17565},
  year={2024},
  url={https://arxiv.org/abs/2406.17565}
}

@article{jin2024p,
  title={{P/D-Serve}: Serving Disaggregated Large Language Model at Scale},
  author={Jin, Yibo and Wang, Tao and Lin, Huimin and Song, Mingyang and Li, Peiyang and Ma, Yipeng and Shan, Yicheng and Yuan, Zhengfan and Li, Cailong and Sun, Yajing and Wu, Tiandeng and Chu, Xing and Huan, Ruizhi and Ma, Li and You, Xiao and Zhou, Wenting and Ye, Yunpeng and Liu, Wen and Xu, Xiangkun and Zhang, Yongsheng and Dong, Tiantian and Zhu, Jiawei and Wang, Zhe and Ju, Xijian and Song, Jianxun and Cheng, Haoliang and Li, Xiaojing and Ding, Jiandong and Guo, Hefei and Zhang, Zhengyong},
  journal={arXiv preprint arXiv:2408.08147},
  year={2024},
  url={https://arxiv.org/abs/2408.08147}
}

@inproceedings{kamath2025pod,
  title={{POD-Attention}: Unlocking Full Prefill-Decode Overlap for Faster {LLM} Inference},
  author={Kamath, Aditya K and Prabhu, Ramya and Mohan, Jayashree and Peter, Simon and Ramjee, Ramachandran and Panwar, Ashish},
  booktitle={Proceedings of the 30th ACM International Conference on Architectural Support for Programming Languages and Operating Systems, Volume 2},
  pages={897--912},
  year={2025},
  url={https://arxiv.org/abs/2410.18038}
}

@article{gao2025duetserve,
  title={{DuetServe}: Harmonizing Prefill and Decode for {LLM} Serving via Adaptive {GPU} Multiplexing},
  author={Gao, Lei and Jiang, Chaoyi and Zarch, Hossein Entezari and Wong, Daniel and Hill, Mark and Annavaram, Murali},
  journal={arXiv preprint arXiv:2511.04791},
  year={2025},
  url={https://arxiv.org/abs/2511.04791}
}

@article{zheng2024sglang,
  title={{SGLang}: Efficient Execution of Structured Language Model Programs},
  author={Zheng, Lianmin and Yin, Liangsheng and Xie, Zhiqiang and Sun, Chuyue and Huang, Jeff and Yu, Cody Hao and Cao, Shiyi and Kozyrakis, Christos and Stoica, Ion and Gonzalez, Joseph E. and Barrett, Clark and Sheng, Ying},
  journal={Advances in Neural Information Processing Systems},
  volume={37},
  pages={62557--62583},
  year={2024},
  url={https://arxiv.org/abs/2312.07104}
}

@inproceedings{yao2025cacheblend,
  title={{CacheBlend}: Fast Large Language Model Serving for {RAG} with Cached Knowledge Fusion},
  author={Yao, Jiayi and Li, Hanchen and Liu, Yuhan and Ray, Siddhant and Cheng, Yihua and Zhang, Qizheng and Du, Kuntai and Lu, Shan and Jiang, Junchen},
  booktitle={Proceedings of the Twentieth European Conference on Computer Systems},
  pages={94--109},
  year={2025},
  url={https://arxiv.org/abs/2405.16444}
}

@inproceedings{gao2024cost,
  title={Cost-Efficient Large Language Model Serving for Multi-turn Conversations with {CachedAttention}},
  author={Gao, Bin and He, Zhuomin and Sharma, Puru and Kang, Qingxuan and Jevdjic, Djordje and Deng, Junbo and Yang, Xingkun and Yu, Zhou and Zuo, Pengfei},
  booktitle={2024 USENIX Annual Technical Conference (USENIX ATC 24)},
  pages={111--126},
  year={2024},
  url={https://www.usenix.org/conference/atc24/presentation/gao-bin-cost}
}

@article{li2025flowkv,
  title={{FlowKV}: A Disaggregated Inference Framework with Low-Latency {KV} Cache Transfer and Load-Aware Scheduling},
  author={Li, Weiqing and Jiang, Guochao and Ding, Xiangyong and Tao, Zhangcheng and Hao, Chuzhan and Xu, Chenfeng and Zhang, Yuewei and Wang, Hao},
  journal={arXiv preprint arXiv:2504.03775},
  year={2025},
  url={https://arxiv.org/abs/2504.03775}
}
\bibliographystyle{mlsys2025}

%%%%%%%%%%%%%%%%%%%%%%%%%%%%%%%%%%%%%%%%%%%%%%%%%%%%%%%%%%%%%%%%%%%%%%%%%%%%%%%
%%%%%%%%%%%%%%%%%%%%%%%%%%%%%%%%%%%%%%%%%%%%%%%%%%%%%%%%%%%%%%%%%%%%%%%%%%%%%%%
% SUPPLEMENTAL CONTENT AS APPENDIX AFTER REFERENCES
%%%%%%%%%%%%%%%%%%%%%%%%%%%%%%%%%%%%%%%%%%%%%%%%%%%%%%%%%%%%%%%%%%%%%%%%%%%%%%%
%%%%%%%%%%%%%%%%%%%%%%%%%%%%%%%%%%%%%%%%%%%%%%%%%%%%%%%%%%%%%%%%%%%%%%%%%%%%%%%
\appendix

%%%%%%%%%%%%%%%%%%%%%%%%%%%%%%%%%%%%%%%%%%%%%%%%%%%%%%%%%%%%%%%%%%%%%%%%%%%%%%%
%%%%%%%%%%%%%%%%%%%%%%%%%%%%%%%%%%%%%%%%%%%%%%%%%%%%%%%%%%%%%%%%%%%%%%%%%%%%%%%

\end{document}